\documentclass[preprint,12pt]{elsarticle}

\usepackage[noend]{algpseudocode}
\usepackage{algorithm}
\usepackage{algorithmicx}
\usepackage{multirow}
\usepackage{microtype}
\usepackage{graphicx}
\usepackage[normalem]{ulem}
\usepackage{makecell}
\useunder{\uline}{\ul}{}
\usepackage{booktabs}
\usepackage{subfigure}
\usepackage{amsmath}
\usepackage{amssymb}
\usepackage{color,xcolor}
\usepackage{url}
\usepackage[hang,flushmargin]{footmisc}

\newcommand{\modelName}{BEER$^2$}
\DeclareMathOperator*{\argmax}{arg\, max}

\journal{Knowledge-Based Systems}

\begin{document}

\begin{frontmatter}



\title{Bidirectional End-to-End Learning of Retriever-Reader Paradigm for Entity Linking}

\author[thu]{Yinghui Li\fnref{cor1}}
\author[damo]{Yong Jiang\fnref{cor1}\corref{cor2}} 
\author[thu]{Yangning Li\fnref{cor1}} 
\author[thu]{Xinyu Lu} 
\author[damo]{Pengjun Xie}
\author[sysu]{Ying Shen}
\author[thu]{Hai-Tao Zheng\corref{cor2}} 

\fntext[cor1]{indicates equal contribution. }
\cortext[cor2]{Corresponding authors.}
\fntext[label3]{E-mails: liyinghu20@mails.tsinghua.edu.cn, yongjiang.jy@alibaba-inc.com, yn-li23@mails.tsinghua.edu.cn, luxy22@mails.tsinghua.edu.cn, chengchen.xpj@taobao.com, zheng.haitao@sz.tsinghua.edu.cn, sheny76@mail.sysu.edu.cn}

\affiliation[thu]{organization={Tsinghua Shenzhen International Graduate School, Tsinghua University},
            city={Shenzhen},
            postcode={518055}, 
            state={Guangdong},
            country={China}}

\affiliation[damo]{organization={Alibaba DAMO Academy},
            city={Hangzhou},
            postcode={311121}, 
            state={Zhejiang},
            country={China}}

\affiliation[sysu]{organization={Sun-Yat Sen University},
            city={Guangzhou},
            postcode={510275}, 
            state={Guangdong},
            country={China}}

\begin{abstract}
Entity Linking (EL) is a fundamental task for Information Extraction and Knowledge Graphs. 
The general form of EL (i.e., end-to-end EL) aims to find mentions in the given document and then link the mentions to corresponding entities in a specific knowledge base.
Recently, the paradigm of retriever-reader promotes the progress of end-to-end EL, benefiting from the advantages of dense entity retrieval and machine reading comprehension.
However, the existing study only trains the retriever and the reader separately in a pipeline manner, which ignores the benefit that the interaction between the retriever and the reader can bring to the task.
To advance the retriever-reader paradigm to perform more perfectly on end-to-end EL, we propose \textbf{\modelName{}}, a \textbf{B}idirectional \textbf{E}nd-to-\textbf{E}nd training framework for \textbf{R}etriever and \textbf{R}eader.
Through our designed bidirectional end-to-end training, \modelName{} guides the retriever and the reader to learn from each other, make progress together, and ultimately improve EL performance.
Extensive experiments on benchmarks of multiple domains demonstrate the effectiveness of our proposed \modelName{}.
\end{abstract}




\end{frontmatter}


\section{Introduction}
\label{Sec:Introduction}

End-to-End Entity Linking (EL)~\citep{6823700, kolitsas-etal-2018-end, DBLP:conf/ijcnlp/ChenLGW20} which is the general form of the EL task, aims to extract mentions from a given text and link the mentions to specific entities in a given knowledge base. Due to its ability to automatically understand text, entity linking has become an essential task for various NLP tasks~\citep{DBLP:journals/corr/abs-2207-08087, DBLP:journals/corr/abs-2305-03688}, such as knowledge graphs construction~\citep{clancy-etal-2019-scalable, DBLP:journals/corr/abs-2302-08774}, automatic text summarization~\citep{amplayo-etal-2018-entity}, and question answering~\citep{DBLP:journals/ibmrd/Ferrucci12}.

\begin{figure*}[h]
\centering
\includegraphics[width=0.80\textwidth]{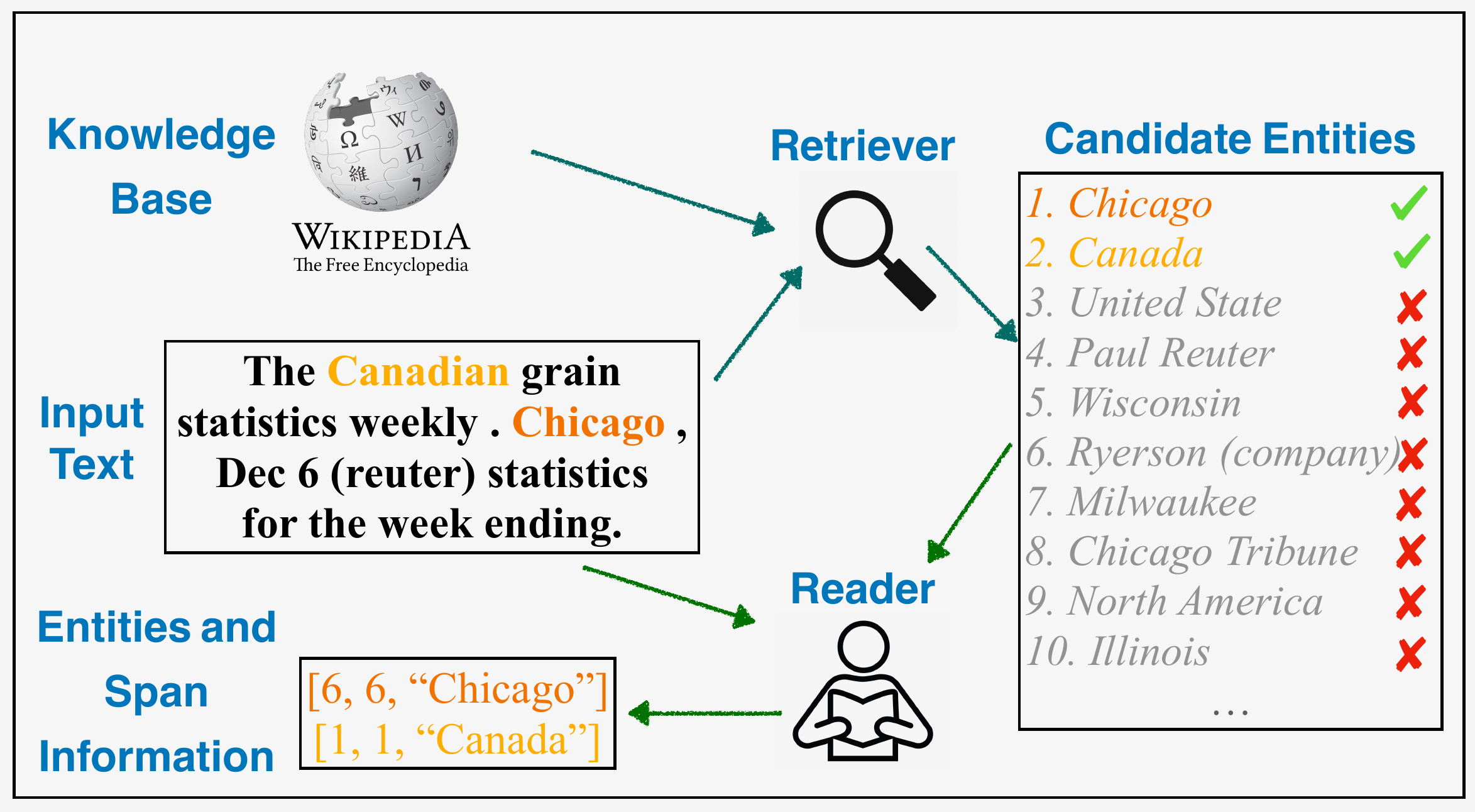}
\caption{An example of the entity linking according to the retriever-reader paradigm.}
\label{Intro_Figure}
\end{figure*}

Early end-to-end EL works~\citep{hoffart-etal-2011-robust, ling-etal-2015-design, luo-etal-2015-joint} mainly divide this task into two subtasks, Mention Detection (MD) and Entity Disambiguation (ED), and study how to exploit the potential relationship between them to improve EL performance.
Most previous works conduct MD before ED~\citep{nguyen-etal-2016-j, martins-etal-2019-joint}, which is an unnatural design because it causes models to predict the span position without entity information.
This is also the essential reason for the long-standing dilemma in end-to-end EL, that is, “MD is more difficult than ED”~\citep{DBLP:conf/aaai/Zhao0ZW19, broscheit-2019-investigating}.
To overcome this challenge, \citep{DBLP:conf/iclr/ZhangHS22} propose EntQA which consists of a retriever-reader structure to perform ED before MD.
As shown in Figure~\ref{Intro_Figure}, given a text, EntQA first retrieves relevant candidate entities from the knowledge base by dense retrieval, and then its reader is responsible for rejecting wrong candidates and extracting the span position information in the text for correct entities.
Benefiting from that dense retrieval effectively reduces the huge search space, EntQA achieves state-of-the-art performance and becomes a strong and advanced baseline of end-to-end EL.

However, EntQA only trains its retriever and reader separately in a pipeline manner, that is, the training of the reader will be performed after the retriever is fully trained. We argue that this pipelined training cannot enable sufficient interaction between the retriever and the reader.
Intuitively, the training signal of the retriever can dynamically affect the training process of the reader, and the results of the reader can also be fed back to the retriever to guide its training. 
Therefore, it is worth studying how to interact between the retriever and reader to improve the end-to-end EL performance.

Motivated by the above intuition, we propose the \textbf{B}idirectional \textbf{E}nd-to-\textbf{E}nd learning of \textbf{R}etriever-\textbf{R}eader (\textbf{\modelName{}}), a more effective training framework that aims to advance the retriever-reader paradigm to perform more perfectly on end-to-end EL.
The \modelName{} contains two data flows in opposite directions: 
(1) Retriever $\rightarrow$ Reader. The retriever dynamically gets candidate entities and inputs them into the reader, thereby updating the training data of the reader in real time.
(2) Reader $\rightarrow$ Retriever. The reader identifies mentions in the documents and inputs the span position information into the retriever, which in turn allows the retriever to perform more effective span-based retrieval.
Through these two bidirectional data flows, we jointly train the retriever and reader in an end-to-end manner and then guide them to learn from each other, make progress together, and finally improve the EL performance.
In addition, we believe that the core idea of our proposed \modelName{} is also useful for enhancing retriever-reader likely models in other tasks, such as open-domain question answering.

In summary, our contributions are in three folds: 
\begin{enumerate}
    \item We present the end-to-end \modelName{} framework, which jointly trains the retriever and reader to make them fully interact and enhance each other.
    \item We conduct extensive experiments on benchmarks in two languages (English and Chinese) of multiple domains (including news, speech, and medical domains) and achieve new state-of-the-art end-to-end EL performance.
    \item We provide sufficient ablation studies and detailed analyses for better verification of the effectiveness of our proposed method.
\end{enumerate}


\begin{figure*}[h]
\centering
\includegraphics[width=1.00\textwidth]{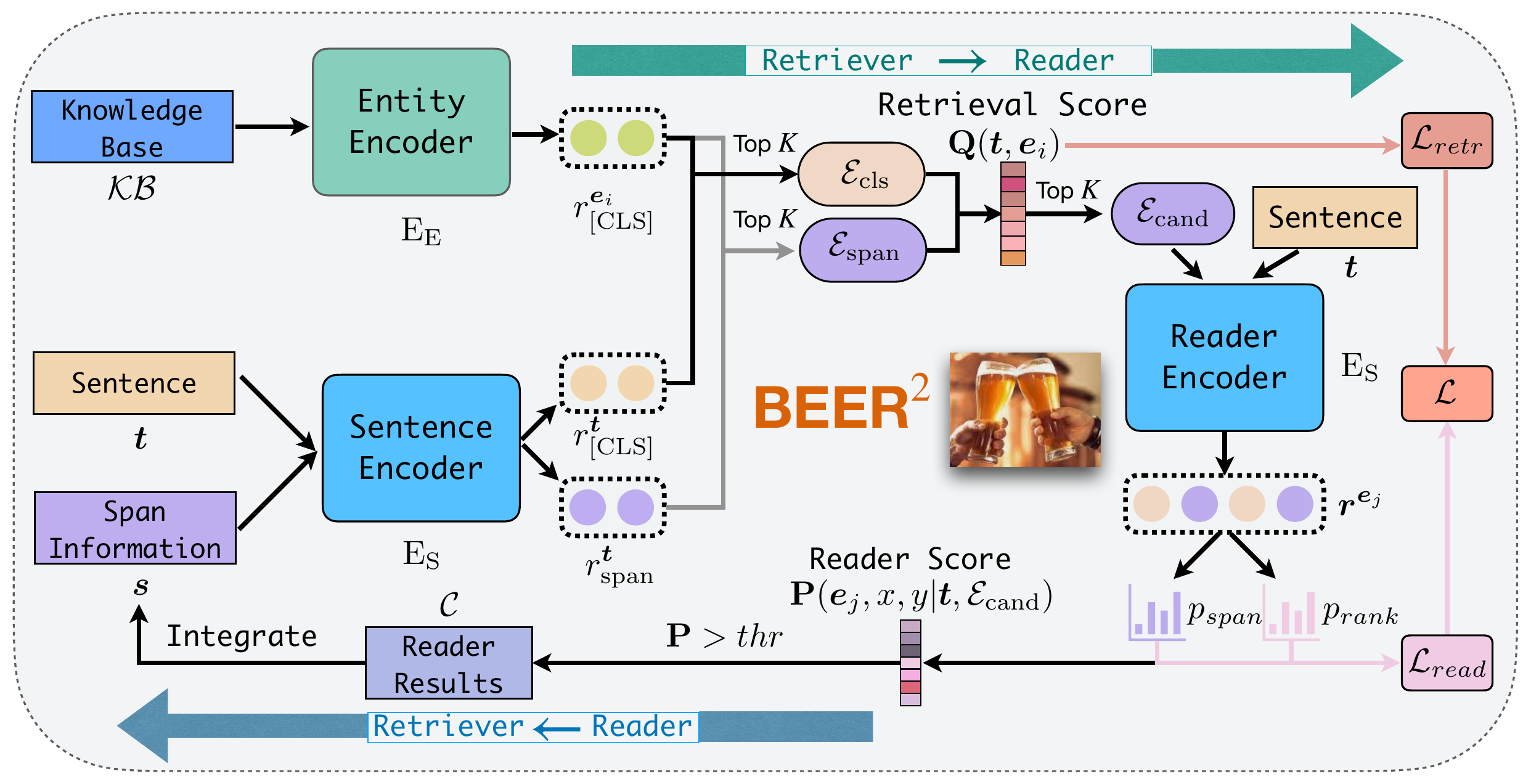}
\caption{The training process of \modelName{}. Note that encoders of the same color represent that they share parameters.}
\label{Method_Figure}
\end{figure*}

\section{Methodology}
\label{Sec:Methodology}
In this section, we introduce the details of \modelName{}, which are illustrated in Figure~\ref{Method_Figure}.
Our proposed \modelName{} consists of a retriever and a reader, which are trained jointly in an end-to-end manner. 
From the perspective of the data flow, our approach contains two opposite data flows, namely “Retriever $\rightarrow$ Reader” and “Reader $\rightarrow$ Retriever”. 
We will first introduce the details of our designed retriever and reader, and then describe the mechanism of the two data flows to explain the training procedure of our method in detail. Finally, we will present the overall training objective of \modelName{}.

\subsection{Retriever-Reader Structure}
\label{sec:retriever-reader-structure}
The retriever module aims to retrieve candidate entities that might belong to the input text from the knowledge base, and the purpose of the reader module is to further reduce the candidate set and predict the specific span position information of the finally predicted entities in the text.

\subsubsection{Retriever}
Let the knowledge base be denoted by $\mathcal{KB} = \{ \boldsymbol{e}_1, ..., \boldsymbol{e}_N \}$. Given a sentence $\boldsymbol{t}$ of length $T_{\boldsymbol{t}}$, the retriever module is to achieve a subset $\mathcal{E}_\text{cand} \subset \mathcal{KB}$ to be the candidate entities of $\boldsymbol{t}$. 
Specifically, we model the retriever as a dual-encoder~\citep{DBLP:conf/nips/BromleyGLSS93}, which contains a sentence encoder $\mathrm{E}_{\text{S}}$ and an entity encoder $\mathrm{E}_{\text{E}}$. We use $\mathrm{E}_{\text{S}}$ to map the sentence $\boldsymbol{t}$ to a sequences of representations $\boldsymbol{r}^{\boldsymbol{t}} $:
\begin{equation}
\begin{aligned}
    \boldsymbol{r}^{\boldsymbol{t}} & =\mathrm{E}_{\text{S}}(\boldsymbol{t}),\\
         & =[r_{\text{[CLS]}}^{\boldsymbol{t}}, r_1^{\boldsymbol{t}}, ..., r_{T_{\boldsymbol{t}}}^{\boldsymbol{t}}],
\end{aligned}
\end{equation}
where $\text{[CLS]}$ is the special token representing the beginning of a sequence in the tokenizer of BERT~\citep{devlin-etal-2019-bert}. And we use $\mathrm{E}_{\text{E}}$ to get the representations $\boldsymbol{r}^{\boldsymbol{e}_i}$ of the entity $\boldsymbol{e}_i \in \mathcal{KB}$:
\begin{equation}
\begin{aligned}
    \boldsymbol{r}^{\boldsymbol{e}_i} & = \mathrm{E}_{\text{E}}(f(\boldsymbol{e}_i)), \\
      & = [r_{\text{[CLS]}}^{\boldsymbol{e}_i}, r_1^{\boldsymbol{e}_i}, ..., r_{T_{\boldsymbol{e}}}^{\boldsymbol{e}_i}],
\end{aligned}
\end{equation}
where $f(\boldsymbol{e}_i)$ represents the operation of obtaining the description text of $\boldsymbol{e}_i$ from $\mathcal{KB}$. It is worth noting that we uniformly limit the description text length of all entities in $\mathcal{KB}$ to $T_{\boldsymbol{e}}$.

After obtaining the representations of the sentence and entities, we select entities based on their retrieval scores. We use the dot product between vectors as our scoring function. We first use the $\text{[CLS]}$ representations of sentences and entities to retrieve the top $K$ entities:
\begin{equation}
\label{equ:ecls}
    \mathcal{E}_\text{cls} = \argmax_{\mathcal{E}^{\prime} \subset \mathcal{E}, \left| \mathcal{E}^{\prime} \right| = K} \sum_{\boldsymbol{e}_i \in \mathcal{E}^{\prime}} r_{\text{[CLS]}}^{\boldsymbol{t}\top}r_{\text{[CLS]}}^{\boldsymbol{e}_i}.
\end{equation}

In addition, to enhance the interaction between the retriever and the reader, we also utilize the span information predicted by the reader for auxiliary retrieval. Assuming that the reader's prediction for the spans in $\boldsymbol{t}$ is $\boldsymbol{s} = \{s_i\}, 1 \leq s_i \leq T_{\boldsymbol{t}}$, we use it to obtain more accurate span representations than $\text{[CLS]}$ representations as follows:
\begin{equation}
\label{equ:rspan}
    r^{\boldsymbol{t}}_\text{span} = \text{avg}([r_{s_i}^{\boldsymbol{t}}]), s_i \in \boldsymbol{s},
\end{equation}
where $\text{avg}(\cdot)$ is the mean pooling operation. Then we use the span representations to retrieve the top $K$ entities again:
\begin{equation}
\label{equ:espan}
    \mathcal{E}_\text{span} = \argmax_{\mathcal{E}^{\prime} \subset \mathcal{E}, \left| \mathcal{E}^{\prime} \right| = K} \sum_{\boldsymbol{e}_i \in \mathcal{E}^{\prime}} r_{\text{span}}^{\boldsymbol{t}\top}r_{\text{[CLS]}}^{\boldsymbol{e}_i}.
\end{equation}

Considering that $\mathcal{E}_\text{cls}$ and $\mathcal{E}_\text{span}$ may have duplicate entities, we finally take the top $K$ entities of the set of  $\{ \mathcal{E}_\text{cls} \cup \mathcal{E}_\text{span} \}$ as the final retrieval result, i.e., $\mathcal{E}_\text{cand}$, which will be sent to the reader as input.

\subsubsection{Reader}
\label{sec:reader}
To enable end-to-end training, we make the reader encoder and the retriever's sentence encoder share parameters. Therefore, we denote the reader encoder as $\mathrm{E}_{\text{S}}$ for the convenience of understanding. Given the output of the retriever (i.e., $\mathcal{E}_\text{cand}$) and the input sentence $\boldsymbol{t}$, for each candidate entity $\boldsymbol{e}_j \in \mathcal{E}_\text{cand}$, we get its joint representation with $\boldsymbol{t}$:
\begin{equation}
\begin{aligned}
    \boldsymbol{r}^{\boldsymbol{e}_j} & = \mathrm{E}_{\text{S}}(\boldsymbol{t} \oplus f(\boldsymbol{e}_j)), \\
     & = [r_{\text{[CLS]}}^{\boldsymbol{e}_j}, r_1^{\boldsymbol{e}_j}, ..., r_{T_{\boldsymbol{t}}}^{\boldsymbol{e}_j}, r_{\text{[SEP]}}^{\boldsymbol{e}_j}, ..., r_{T_{\boldsymbol{t}}+T_{\boldsymbol{e}}}^{\boldsymbol{e}_j}].
\end{aligned}
\end{equation}
Based on the joint representation, according to the mechanism proposed in EntQA, we compute the probability of span $(x, y), 1 \leq x, y \leq T_{\boldsymbol{t}}$ and the ranking probability of $\boldsymbol{e}_j, 1 \leq j \leq K$:
\begin{gather}
    p_{1}(x|\boldsymbol{t}, \boldsymbol{e}_j) = \operatorname{softmax}(\boldsymbol{W}_{1} \boldsymbol{r}^{\boldsymbol{e}_j})[x] , \\
    p_{2}(y|\boldsymbol{t}, \boldsymbol{e}_j) = \operatorname{softmax}(\boldsymbol{W}_{2} \boldsymbol{r}^{\boldsymbol{e}_j})[y] , \\
    p_{span}(x,y|\boldsymbol{t}, \boldsymbol{e}_j) = p_{1}(x|\boldsymbol{t}, \boldsymbol{e}_j) \times p_{2}(y|\boldsymbol{t}, \boldsymbol{e}_j), \\
    p_{rank}(\boldsymbol{e}_j|\boldsymbol{t}, \mathcal{E}_\text{cand}) = \frac{\exp \left(\boldsymbol{W}_{3}^{\top} r_1^{\boldsymbol{e}_j}\right)}{\sum_{j^{\prime}=1}^K \exp \left(\boldsymbol{W}_{3}^{\top} r_1^{{\boldsymbol{e}_{j^\prime}}}\right)} ,
\end{gather}
where $\boldsymbol{W}_{1}, \boldsymbol{W}_{2}, \boldsymbol{W}_{3}$ are trainable parameters.
Furthermore, for a combination of span and entity, its reader score is computed as:
\begin{equation}
\begin{aligned}
    \mathbf{P}(\boldsymbol{e}_j, x, y |\boldsymbol{t}, \mathcal{E}_\text{cand} ) = p_{span}(x,y|\boldsymbol{t}, \boldsymbol{e}_j) \\ \times p_{rank}(\boldsymbol{e}_j|\boldsymbol{t}, \mathcal{E}_\text{cand}).
\end{aligned}
\end{equation}

Based on the scores of all possible combinations, we select the combinations that satisfy “$\mathbf{P} > thr$” as the final reader prediction results. The $thr$ is the threshold we choose empirically. We denote the $N$ reader results as $\mathcal{C} = \{ (\boldsymbol{e}_l, x_l, y_l ) \}, 1 \leq l \leq N$. We integrate the span position (i.e., $\{ (x_l, y_l)\}$) into $\boldsymbol{s} = \{s_i\}$ and send $\boldsymbol{s}$ to the retriever.

\subsection{Bidirectional Data Flows}
\textbf{The key innovation of \modelName{} compared to EntQA is to use two bidirectional data flows so that the retriever and the reader are trained in an end-to-end fashion.} Their enhanced interaction allows them to obtain positive affect from each other and improve performance.

\subsubsection{Retriever $\rightarrow$ Reader }
\label{sec:retriever_reader}
In the framework of \modelName{}, the retriever sends its retrieval results $\mathcal{E}_\text{cand}$ to the reader after each time it completes the retrieval. 
In fact, in EntQA, the input of the reader is also the output of the retriever. However, the pipeline training method of EntQA determines that only when the retriever's training ends, the inference result of the retriever module can be used as the training input of the reader. 
We think this is a kind of data interaction that is relatively hard and dull, and the reader cannot perceive the signal change of the retriever training process, because the reader can only receive the inference result of the retriever. 
Therefore, unlike EntQA, \modelName{} dynamically sends the retriever results to the reader during the training process, which allows the reader to learn the experience gained by the retriever when they are training. Additionally, the end-to-end property of the \modelName{} framework enables the retriever and reader to receive gradient propagation at the same time during training. If the retriever selects the wrong candidate entity, this will cause the reader to be affected as well, and the signal of the reader will also be directly fed back to the retriever so that it can correct the error in time. Particularly, we believe that if the signal of the ranking probability $p_{rank}$ calculated in the reader is propagated to the retriever, it will be very helpful for the optimization of the retriever, because the ranking probability is to score and sort the results of the retriever, so this can obviously be regarded as the training rewards of the retriever.

\subsubsection{Reader $\rightarrow$ Retriever}
\label{sec:reader_retriever}
From Equations~\ref{equ:rspan} and~\ref{equ:espan}, we know the key information connecting the bridge from the reader to the retriever is the span position information $\boldsymbol{s}$. According to the span prediction results of the reader, our retriever accurately extracts the span representations of the input sentence, and then performs auxiliary entity retrieval. There are two main motivations for us to use the span prediction results to assist retrieval: 
(1) The $\text{[CLS]}$ representation of a sentence can reflect the semantics of the entire sentence, but it is not sufficient for the representation of the entities in the sentence. Hence, we think only using $\text{[CLS]}$ representations is not optimal for entity-centric tasks~\citep{DBLP:conf/sigir/LiLHYS022, zaporojets2022tempel, DBLP:journals/corr/abs-2212-01612} like EL. 
The span representations can not only make the retriever perceive the location of the mentions in the sentence, but also improve the retrieval diversity. 
(2) Also benefiting from our parameter sharing setting, during training, if the reader makes a wrong span prediction, then this will cause the retriever to fail to obtain an accurate span representation and make a wrong entity selection, and the retriever's loss gradient will be passed back to the reader so that it can learn and get progress.

Besides, in practice, to allow the retriever to have span information input at the beginning of training, we arrange a process similar to the model's warm-up before starting formal training. This process will sequentially pre-train the retriever and reader with a small number of epochs, so as to obtain the span information which will be used for the initial input of the retriever in the formal end-to-end training of \modelName{}. It is worth noting that this warm-up-like process does not cause an unfair comparison between our method and EntQA, which we will empirically prove in Section~\ref{exp:effects_epoch}.

\subsection{Overall Training Objective}
In Section~\ref{sec:retriever-reader-structure},  we have introduced the working mechanism of our retriever and reader in detail (this is also the details of their inference process), now we introduce the training objective of \modelName{}.

Given a training sentence $\boldsymbol{t}$ and the knowledge base $\mathcal{KB}$, based on the Equations~\ref{equ:ecls} and~\ref{equ:espan}, for an entity $\boldsymbol{e}_i \in \mathcal{KB}$, the retriever's score is defined as:
\begin{equation}
    \mathbf{Q}(\boldsymbol{t}, \boldsymbol{e}_i) = r_{\text{[CLS]}}^{\boldsymbol{t}\top}r_{\text{[CLS]}}^{\boldsymbol{e}_i} + r_{\text{span}}^{\boldsymbol{t}\top}r_{\text{[CLS]}}^{\boldsymbol{e}_i}.
\end{equation}

For the training sentence $\boldsymbol{t}$, we have its gold entity set $\mathcal{G} \subset \mathcal{KB}$, and we can achieve its negative entity set $\mathcal{N} \subset \mathcal{KB} \setminus \mathcal{G}$. Then we train the retriever by Noise Contrastive Estimation objective~\citep{DBLP:journals/jmlr/GutmannH10} which is defined as: 
\begin{equation}
\begin{aligned}
    & \mathcal{L}_{retr} = \max \sum_{\boldsymbol{g} \in \mathcal{G}} \log \\ 
    & \left( \frac{\exp \left(\mathbf{Q}(\boldsymbol{t}, \boldsymbol{g})\right)}{\exp \left(\mathbf{Q}(\boldsymbol{t}, \boldsymbol{g})\right)+\sum_{\boldsymbol{n} \in \mathbf{N}} \exp \left(\mathbf{Q}\left(\boldsymbol{t}, \boldsymbol{n}\right)\right)}\right).
\end{aligned}
\end{equation}

As for the training of the reader, we directly optimize it to maximize the span probability (i.e., $p_{span}$) and ranking probability (i.e., $p_{rank}$):
\begin{equation}
\begin{aligned}
    \mathcal{L}_{read} = \max \sum_{j=1}^{K} \sum_{(x, y)} ( \log p_{span}(x,y|\boldsymbol{t}, \boldsymbol{e}_j) + \\  \log p_{rank}(\boldsymbol{e}_j|\boldsymbol{t}, \mathcal{E}_\text{cand})),
\end{aligned}
\end{equation}
It is worth noting that, during the training process, the combination of the $(x,y)$ is the gold mention spans of every candidate entity $\boldsymbol{e}_j \subset \mathcal{E}_\text{cand}$.

Finally, we train the retriever objective $\mathcal{L}_{retr}$ and reader objective $\mathcal{L}_{read}$ simultaneously. The overall end-to-end objective of \modelName{} is defined as:
\begin{equation}
    \mathcal{L} = \mathcal{L}_{retr} + \mathcal{L}_{read}.
\end{equation}


\section{Experiments}
\label{Sec:Experiments}

\begin{table}[h]
\tiny
\centering
\begin{tabular}{lrrrr}
\hline \textbf{Dataset} & \textbf{\#Train} & \textbf{\#Dev} & \textbf{\#Test} & \textbf{\#KB} \\
\hline AIDA-CoNLL & 946 & 216 & 231 & 5,903,530 \\
BC5CDR & 500 & 500 & 500 & 2,311 \\
NLPCC2022 & 1,549 & 387 & 387 & 118,795 \\
CCKS2020 & 70,000 & 10,000 & 10,000 & 360,000 \\
\hline
\end{tabular}

\caption{Statistics of the datasets that we use. \#Train/\#Dev/\#Test represents the number of documents in Training/Development/Test sets respectively, and \#KB represents the number of entities in the knowledge base used by the corresponding dataset.}
\label{Data_Statistics}
\end{table}

\subsection{Datasets}
To evaluate \modelName{} comprehensively, we select EL datasets from multiple domains to conduct experiments. Particularly, it is well known that the medical domain is very different from other domains due to its professional nature, so medical EL has long been regarded as an independent task~\citep{mondal-etal-2019-medical}. But we run \modelName{} on both the medical domain and other generic domains. Additionally, our work is also the first to use multilingual benchmarks, including English and Chinese. The dataset statistics are shown in Table~\ref{Data_Statistics}.
\begin{itemize}
    \item \textbf{News:} The AIDA-CoNLL dataset~\citep{hoffart-etal-2011-robust} contains 1,393 English news articles from Reuters. Its entities are identified by YAGO2 entity name and Wikipedia URL. 
    Following previous works~\citep{DBLP:conf/iclr/CaoI0P21, DBLP:conf/iclr/ZhangHS22}, we split the AIDA-CoNLL dataset into training (946 documents), development (216 documents), and test (231 documents) sets. We use KILT~\citep{petroni-etal-2021-kilt} which contains 5,903,530 entities as the given knowledge base. 
    \item \textbf{Medical:} The BC5CDR dataset~\citep{DBLP:journals/biodb/LiSJSWLDMWL16} contains 1,500 medical abstracts that are annotated with MeSH ontology. 
    Same as related works~\citep{DBLP:conf/aaai/Zhao0ZW19, zhou-etal-2021-end}, we equally divide it into training, development, and test sets. Because the BC5CDR corpus is annotated with MeSH ontology, we also use MeSH which has 2,311 medical concepts as the knowledge base. 
    \item \textbf{Speech:} The NLPCC2022 dataset~\citep{DBLP:conf/nlpcc/SongZTG22} is the benchmark for the public competition of the conference of NLPCC2022. It consists of 1,936 TED talks converted from raw audio. 
    We manually and randomly split the full dataset into 4:1 for training and testing. Besides, the competition organizers officially provide a knowledge base constructed based on Wikidata. This knowledge base contains 118,795 entities.
    \item \textbf{Short-Text:} The CCKS2020 dataset~\footnote{\url{http://biendata.xyz/competition/ccks_2020_el}} is provided by CCKS2020 Chinese short text EL task. Its corpus comes from various domains, such as movies, TV, novels, etc. 
    The training data includes 70,000 sentences and the validation/test includes 10,000 sentences. The given knowledge base is from the Baidu Baike and includes approximately 360,000 entities.
\end{itemize}

\subsection{Baseline Methods}
To reflect the competitiveness of \modelName{}, we select several advanced strong baselines:
\textbf{End2End EL}~\citep{kolitsas-etal-2018-end} uses a Bi-LSTM~\citep{DBLP:journals/neco/HochreiterS97} to encode embeddings and links mention to entities based on local and global scores.
\textbf{Joint NER EL}~\citep{martins-etal-2019-joint} propose to jointly learn the NER task and EL task to make them benefit from each other.
\textbf{REL}~\citep{REL} is a widely used open-source toolkit for entity linking. It is an ensemble of multiple state-of-the-art NLP methods and packages.
\textbf{GENRE}~\citep{DBLP:conf/iclr/CaoI0P21} models the EL task as a seq2seq problem and automatically generates unique entity identifiers of the input guiding text.
\textbf{ReFinED}~\citep{ayoola-etal-2022-refined} is an efficient zero-shot-capable method for end-to-end EL. It introduces the fine-grained entity typing task to improve the performance of EL.
\textbf{EntQA~\footnote{\url{https://github.com/WenzhengZhang/EntQA}}}~\citep{DBLP:conf/iclr/ZhangHS22} decomposes the end-to-end EL task into two subproblems, namely entity retrieval and question answering. EntQA is the previous state-of-the-art method on the AIDA-CoNLL dataset.
\textbf{DNorm}~\citep{Dnorm} utilizes the TF-IDF to learn a bilinear mapping function for ED of the medical EL task.
\textbf{IDCNN}~\citep{strubell-etal-2017-fast} is a deep learning-based method that uses a CNN network to do the medical NER task.
\textbf{E2EMERN}~\citep{zhou-etal-2021-end} is an end-to-end progressive multi-task learning framework for the medical EL task. It achieves the previous state-of-the-art results on the BC5CDR dataset.
\textbf{KENER}~\citep{KENER} focuses on incorporating proper knowledge in the MD subtask to improve the overall performance of linking. It is the 1st system in the competition on NLPCC2022.

\subsection{Evaluation Metric}
To ensure the fairness of the comparison between our method and existing models, the metrics we use to report our main results are the widely used InKB Micro Precision, Recall, and F1 score. Specifically, for the end-to-end EL task we are concerned about, a mention is considered to be correct only when its span position is extracted correctly and the corresponding entity id in the knowledge base is predicted correctly. It should be emphasized that when evaluating EL performance, the F1 score is considered as the primary metric. The F1 score is the harmonic mean of Precision and Recall, meaning it takes both of them into account when calculating overall performance and effectively balances the trade-off between Precision and Recall. A high F1 score indicates that model performs well in terms of correctly identifying and linking entities while minimizing false positives and false negatives. 
In addition, because both EntQA and \modelName{} are models of the retriever-reader paradigm, we report the retriever's Recall@K to reflect the retrieval performance.

\subsection{Implementation Details}
All the source codes of our experiments are implemented using Pytorch~\citep{DBLP:conf/nips/PaszkeGMLBCKLGA19}.
The architectures of the encoders (i.e., retrieval dual-encoder and reader encoder) we use are  $\text{BERT}_{\text{LARGE}}$-like models.  For different domains, we use different backbone parameters to initialize the base encoders. For the news and speech domains, we initialize encoders with pre-trained BLINK~\citep{wu-etal-2020-scalable}. For the initial parameter of encoders of the medical domain and Chinese language, we select Biobert-Large~\citep{BioBERT} and Chinese-Roberta-Wwm-Ext~\citep{Chinese-Roberta-Wwm-Ext}, respectively. For the selection of different base backbone models, \modelName{} achieves stable and competitive performance improvements, which reflects the generalization and effectiveness of our proposed method. We train \modelName{} with the Adam~\citep{Adam} optimizer for 10 epochs. And our model is trained with linear decay and learning rate warming up. The initial retriever learning rate is set to 2e-6 and the initial reader learning rate is set to 1e-5. The training batch size is set to 8 and the evaluation batch size is set to 32. The number of retrieved candidate entities $K$ is set to 120 by default. We choose the default threshold parameter $thr$ as 0.03. 
We break up each document into sentences of $T_{\boldsymbol{t}} = 32$ and pad the description text of entities in the knowledge base to $T_{\boldsymbol{t}} = 128$. We use Faiss~\footnote{\url{https://github.com/facebookresearch/faiss}}~\citep{Faiss} for fast entity retrieval. 

\begin{table*}[ht]
\tiny
\centering
\begin{tabular}{@{}c|c|c|l|p{1.5cm}p{1.5cm}p{1.5cm}@{}}
\toprule
\multirow{2}{*}{\textbf{Language}} & \multirow{2}{*}{\textbf{Domain}} &  \multirow{2}{*}{\textbf{Dataset}} & \multicolumn{1}{c|}{\multirow{2}{*}{\textbf{Method}}} & \multicolumn{3}{c}{\textbf{InKB Micro}} \\
 \multicolumn{1}{c|}{} & \multicolumn{1}{c|}{} & \multicolumn{1}{c|}{} & \multicolumn{1}{c|}{} & \textbf{Precision} & \textbf{Recall} & \textbf{F1 Score}  \\ \midrule
 
\multicolumn{1}{c|}{\multirow{7}{*}{\textbf{English}}} & \multicolumn{1}{c|}{\multirow{7}{*}{\textbf{News}}} & \multicolumn{1}{c|}{\multirow{7}{*}{AIDA-CoNLL}} 
& End2End EL~\citep{kolitsas-etal-2018-end}  & 80.9 & 84.0 & 82.4  \\
\multicolumn{1}{l|}{} & \multicolumn{1}{l|}{} & \multicolumn{1}{l|}{} & Joint NER EL~\citep{martins-etal-2019-joint}  & 81.1 & 82.8 & 81.9  \\
\multicolumn{1}{l|}{} & \multicolumn{1}{l|}{} & \multicolumn{1}{l|}{} & REL~\citep{REL}  & 79.5 & 81.5 & 80.5 \\
\multicolumn{1}{l|}{} & \multicolumn{1}{l|}{} & \multicolumn{1}{l|}{} & GENRE~\citep{DBLP:conf/iclr/CaoI0P21}  & 81.7 & 85.8 & 83.7\\
\multicolumn{1}{l|}{} & \multicolumn{1}{l|}{} & \multicolumn{1}{l|}{} & ReFinED~\citep{ayoola-etal-2022-refined}  & 81.8 & 86.3 & 84.0\\
\cmidrule(l){4-7} 
\multicolumn{1}{l|}{} & \multicolumn{1}{l|}{} & \multicolumn{1}{l|}{} & EntQA~\citep{DBLP:conf/iclr/ZhangHS22}  & \underline{84.6} & \underline{87.0} & \underline{85.8} \\
\multicolumn{1}{l|}{} & \multicolumn{1}{l|}{} & \multicolumn{1}{l|}{} & \modelName{} (Ours) & \textbf{86.9}$^\uparrow$ & \textbf{87.2}$^\uparrow$ & \textbf{87.0}$^\uparrow$ \\

\midrule

\multicolumn{1}{c|}{\multirow{5}{*}{\textbf{English}}} & \multicolumn{1}{c|}{\multirow{5}{*}{\textbf{Medical}}} & \multicolumn{1}{c|}{\multirow{5}{*}{BC5CDR}} 
& DNorm~\citep{Dnorm}  & \underline{82.7} & 78.7 & 80.7  \\
\multicolumn{1}{l|}{} & \multicolumn{1}{l|}{} & \multicolumn{1}{l|}{} & IDCNN~\citep{strubell-etal-2017-fast} & 82.0 & 80.3 & 81.1 \\
\multicolumn{1}{l|}{} & \multicolumn{1}{l|}{} & \multicolumn{1}{l|}{} & E2EMERN~\citep{zhou-etal-2021-end} & 82.5 & \underline{\textbf{82.1}} & \underline{82.3}\\ 
\cmidrule(l){4-7} 
\multicolumn{1}{l|}{} & \multicolumn{1}{l|}{} & \multicolumn{1}{l|}{} & EntQA~\citep{DBLP:conf/iclr/ZhangHS22} & 81.8 & 81.2 & 81.5 \\
\multicolumn{1}{l|}{} & \multicolumn{1}{l|}{} & \multicolumn{1}{l|}{} & \modelName{}~(Ours) & \textbf{86.0}$^\uparrow$ & 80.3 & \textbf{83.1}$^\uparrow$  \\ 

\midrule

\multicolumn{1}{c|}{\multirow{3}{*}{\textbf{English}}} & \multicolumn{1}{c|}{\multirow{3}{*}{\textbf{Speech}}} & \multicolumn{1}{c|}{\multirow{3}{*}{NLPCC2022}} & KENER~\citep{KENER}  & - & - & \underline{74.6}  \\
\cmidrule(l){4-7} 
\multicolumn{1}{l|}{} & \multicolumn{1}{l|}{} & \multicolumn{1}{l|}{} & EntQA~\citep{DBLP:conf/iclr/ZhangHS22} & \underline{76.0} & \underline{72.1} & 74.0 \\
\multicolumn{1}{l|}{} & \multicolumn{1}{l|}{} & \multicolumn{1}{l|}{} & \modelName{}~(Ours) & \textbf{76.1}$^\uparrow$ & \textbf{77.5}$^\uparrow$ & \textbf{76.8}$^\uparrow$  \\ 
\midrule

\multicolumn{1}{c|}{\multirow{2}{*}{\textbf{Chinese}}} & \multicolumn{1}{c|}{\multirow{2}{*}{\textbf{Short-Text}}} & \multicolumn{1}{c|}{\multirow{2}{*}{CCKS2020}} & EntQA~\citep{DBLP:conf/iclr/ZhangHS22}  & \underline{75.5} & \underline{70.7} & \underline{73.0}  \\
\multicolumn{1}{l|}{} & \multicolumn{1}{l|}{} & \multicolumn{1}{l|}{} & \modelName{}~(Ours) & \textbf{77.1}$^\uparrow$ & \textbf{72.8}$^\uparrow$ & \textbf{74.9}$^\uparrow$ \\

\bottomrule
\end{tabular}

\caption{The performance of \modelName{} and all baselines. We underline the previous state-of-the-art results for convenient comparison. Note that the result of E2EMERN is obtained by running its officially trained model under our metric.
}
\label{Main_Results}
\end{table*}

\subsection{Experimental Results}
From Table~\ref{Main_Results}, we know that:
\begin{enumerate}
    \item Our \modelName{} outperforms the previous state-of-the-art models on all datasets. Specifically, for the main metric, \modelName{} exceeds EntQA by 1.2\% F1 on AIDA-CoNLL, exceeds E2EMERN by 0.8\% F1 on BC5CDR, exceeds KENER by 2.2\% F1 on NLPCC2022, exceeds EntQA by 1.9\% F1 on CCKS2020. The strong results on multiple domains obtained by \modelName{} demonstrate the effectiveness of our proposed bidirectional end-to-end learning of the retriever-reader paradigm for EL.
    \item Compared with EntQA, the improvements of \modelName{} are significant. For the domains of medicine and speech, in which the performance of EntQA is not the best, \modelName{} outperforms it by 2.6\% F1 and 2.8\% F1 respectively and becomes the best model in these two domains. This indicates that \modelName{} has better domain adaptation ability.
    \item For the speech domain, \modelName{} outperforms KENER, which reflects the competitiveness of \modelName{}. KENER is a competition system that includes an ensemble method. Without this trick that is useful in improving performance, \modelName{} is still better than KENER. Additionally, the better performance compared with EntQA on the Chinese benchmark also reflects the language robustness of \modelName{}.
\end{enumerate}

\begin{figure*}[h]
\centering
\includegraphics[width=0.60\textwidth]{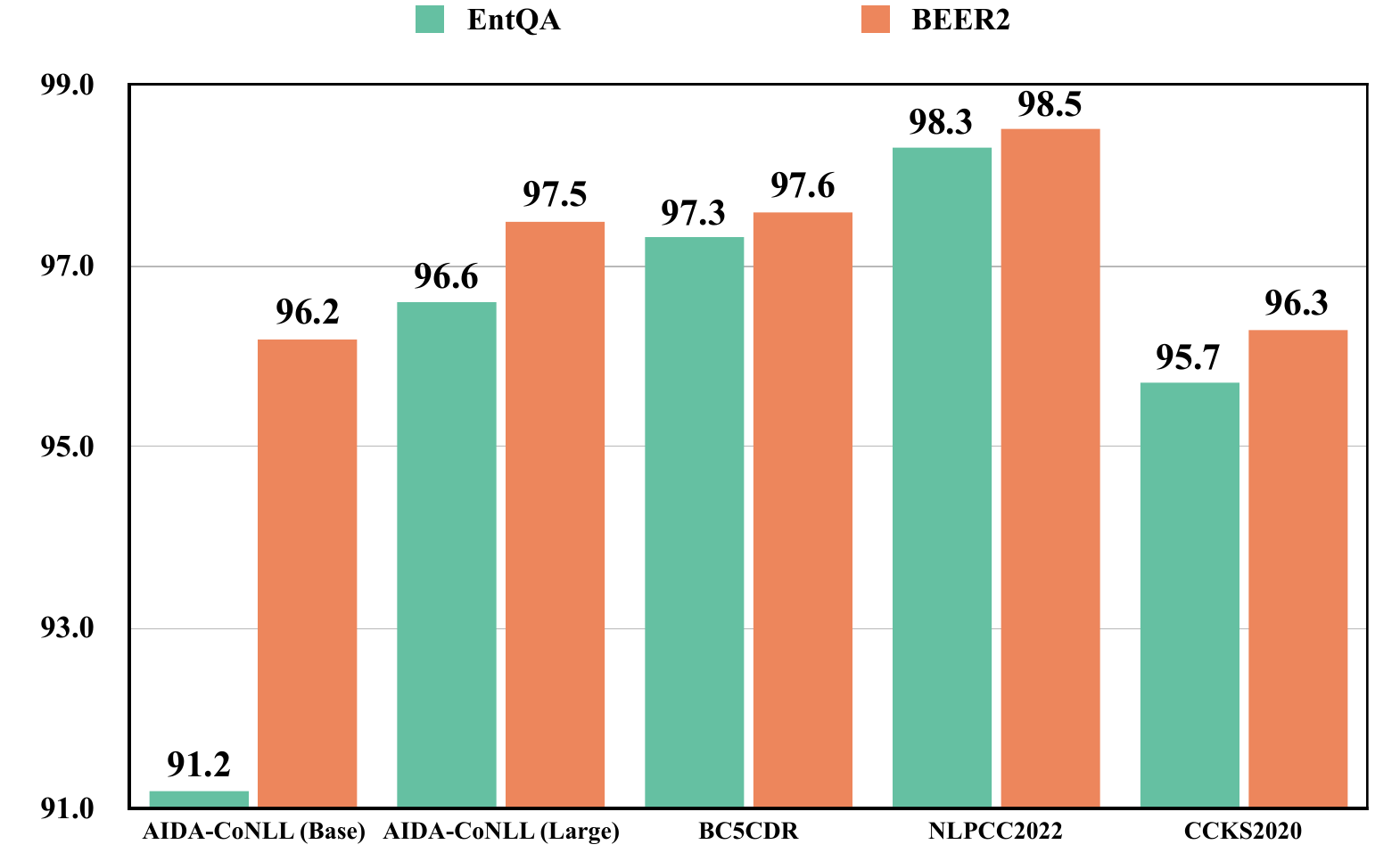}
\caption{The retrieval performance (Recall@K) of EntQA and \modelName{}. Particularly, on AIDA-CoNLL, we use BERT-Base/Large to initialize respectively.}
\label{Retriever_Performance_Figure}
\end{figure*}

\subsection{Analysis and Discussion}
\subsubsection{The Retrieval Performance}
Table~\ref{Main_Results} reports the reader performance as our main results. As a study of the retriever-reader structure, it is necessary to analyze the retrieval performance. From Figure~\ref{Retriever_Performance_Figure}, we see that \modelName{} always has better retrieval performance than EntQA, which verifies the advantage of our designed retriever module. When Bert-Base is used as the backbone, the result of EntQA is relatively low, which leaves more space for \modelName{}. Therefore, our method indeed improves significantly. Furthermore, we also find that the improvements on AIDA-CoNLL and CCKS2020 are greater than that on BC5CDR and NLPCC2022. For this phenomenon, we suspect that it is because the knowledge bases of AIDA-CoNLL and CCKS2020 have more entities than BC5CDR and NLPCC2022. A larger number of entities pose a greater challenge to the retriever, thus, better performance on larger knowledge bases reflects that our retriever is better than EntQA's.

\begin{table}[t]
\tiny
\centering
\begin{tabular}{@{}l|cc@{}}
\toprule
\multicolumn{1}{c|}{\textbf{Method}}  & \textbf{Recall@K} & \textbf{F1} \\
\midrule
\noindent EntQA (Base)   & 91.2 & 79.1\\
\modelName{} (Base, Retriever $\rightarrow$ Reader) & 93.1 & 80.3 \\
\modelName{} (Base, Reader $\rightarrow$ Retriever) & 95.8 & 81.4 \\
\modelName{} (Base, Retriever $\leftrightarrow$ Reader) & 96.2 & 81.7 \\
\midrule
\noindent EntQA (Large)   & 96.6 & 85.8 \\
\modelName{} (Large, Retriever $\rightarrow$ Reader) & 96.8 & 86.0 \\
\modelName{} (Large, Reader $\rightarrow$ Retriever) & 97.2 & 86.3 \\
\modelName{} (Large, Retriever $\leftrightarrow$ Reader) & 97.5 & 87.0 \\
\midrule
\noindent \modelName{}  & \textbf{97.5} & \textbf{87.0}  \\
\bottomrule
\end{tabular}

\caption{The retriever and reader performance of the variants of \modelName{} on AIDA-CoNLL. }
\label{TwoFlows_Performance_Table}
\end{table}

\subsubsection{Effects of Two Data Flows}
Our technical contribution is that we design an end-to-end training mechanism, which includes two bidirectional data flows as bridges connecting the retriever and reader modules. Therefore, we further conduct ablation studies on these two data flows. 

From Table~\ref{TwoFlows_Performance_Table}, we see that each of the data flows we design individually brings considerable improvements. As described in Sections~\ref{sec:retriever_reader} and~\ref{sec:reader_retriever}, while dynamically inputting the candidate entities from the retriever effectively helps the reader's training, thanks to the end-to-end training, the retriever itself is also further optimized. This view can be seen from the results that \modelName{} (Retriever $\rightarrow$ Reader) is better than EntQA on both Recall@K and F1. Similarly, from the comparison of the results of \modelName{} (Reader $\rightarrow$ Retriever) and EntQA, it can be known that the span information sent from the reader to the retriever not only effectively assists the work of the retriever, but also makes its own progress in the prediction of the span position. Besides, the results of \modelName{} (Retriever $\leftrightarrow$ Reader) show that these two data flows cooperate well in the framework of \modelName{}, resulting in better performance than they obtain alone.

\begin{table}[t]
\tiny
\centering
\begin{tabular}{@{}l|ccc@{}}
\toprule
\multicolumn{1}{c|}{\textbf{Method}}  & \textbf{Pre} & \textbf{Rec} & \textbf{F1} \\
\midrule
\noindent EntQA (4 epochs)   & 84.6 & 87.0 & 85.8 \\
\noindent EntQA (20 epochs)   & 85.4 & 86.4 & 85.9 \\
\modelName{} (1 epoch + 10 epochs) & 86.6 & 87.4 & 87.0 \\
\modelName{} (5 epochs + 10 epochs) & 86.9 & 87.2 & 87.0 \\
\bottomrule
\end{tabular}

\caption{The performance when training models with different epochs on AIDA-CoNLL. }
\label{Epochs_Effect_Table}
\end{table}

\subsubsection{Effects of Training Epochs}
\label{exp:effects_epoch}
To verify that the warm-up-like pre-training process before the formal training starts does not cause unfairness in model comparison, we train EntQA and \modelName{} for different epochs.
Note that \modelName{} (1 epoch + 10 epochs) of Table~\ref{Epochs_Effect_Table} means that we pre-train the retriever and reader for 1 epoch and then formally train them for 10 epochs under the end-to-end setting. 
From Table~\ref{Epochs_Effect_Table}, even EntQA is trained for 20 epochs, which means that the number of training epochs is more than that of \modelName{}, its performance is only slightly improved compared to EntQA (4 epochs). This shows that simply increasing the number of training epochs does not bring substantial performance improvements for EntQA. 
Moreover, the slight difference in results between \modelName{} (1 epoch + 10 epochs) and \modelName{} (5 epochs + 10 epochs) also indicates that the number of epochs of the warm-up-like process is not critical to the training of \modelName{}. The role of this process is only to provide initial span information for our designed retriever. The great advantage of \modelName{} (5 epochs + 10 epochs) compared to EntQA (20 epochs) also empirically proves what we mentioned in Section~\ref{sec:reader_retriever}, that is, the warm-up-like pre-training process will not cause unfairness in the comparison between EntQA and \modelName{}.

\begin{figure}[ht]
\centering
\subfigure[$K$ changes] { \label{Parameters_K_Figure} 
\includegraphics[scale=0.45]{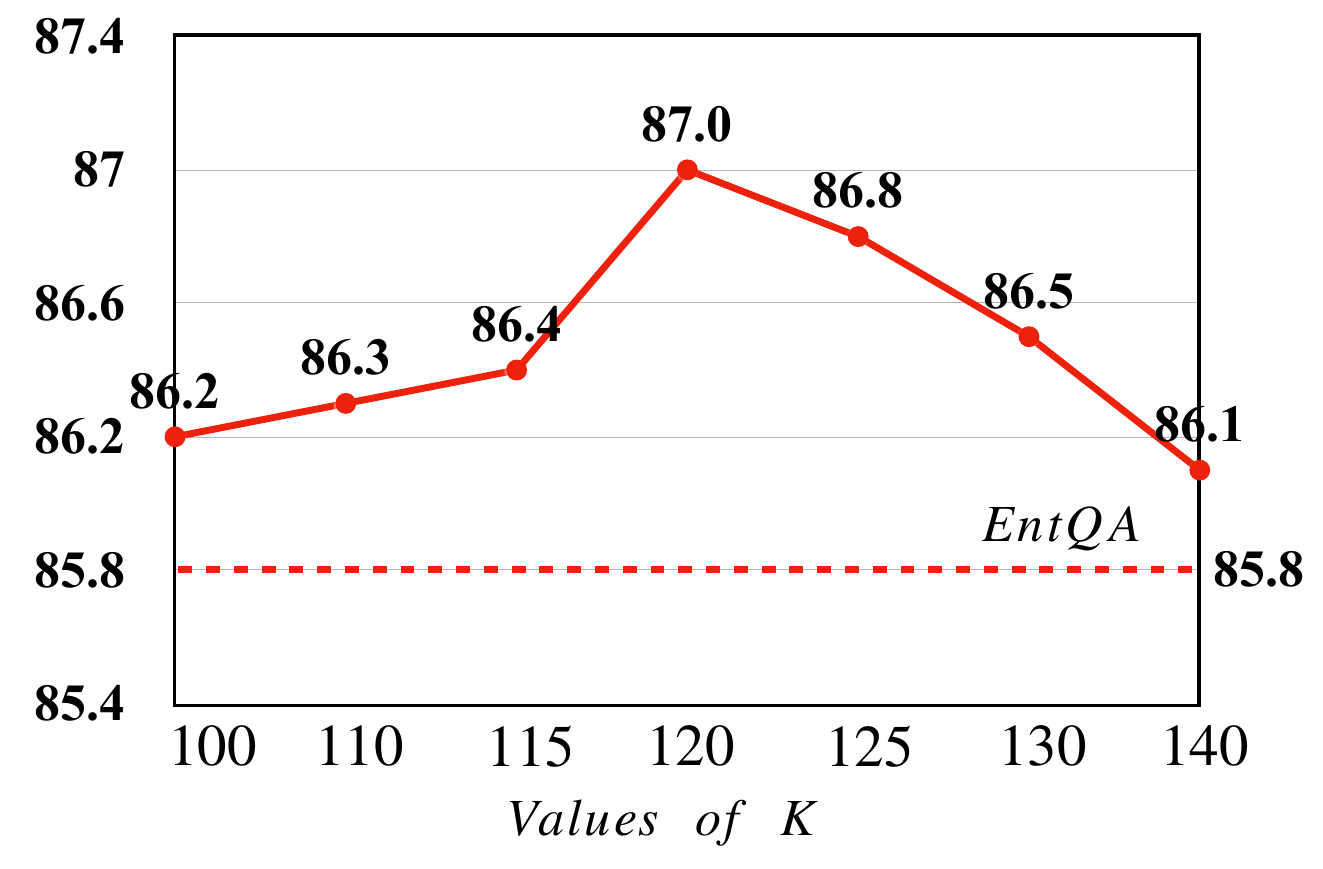} 
} 
\subfigure[$thr$ changes] { \label{Parameters_thr_Figure} 
\includegraphics[scale=0.45]{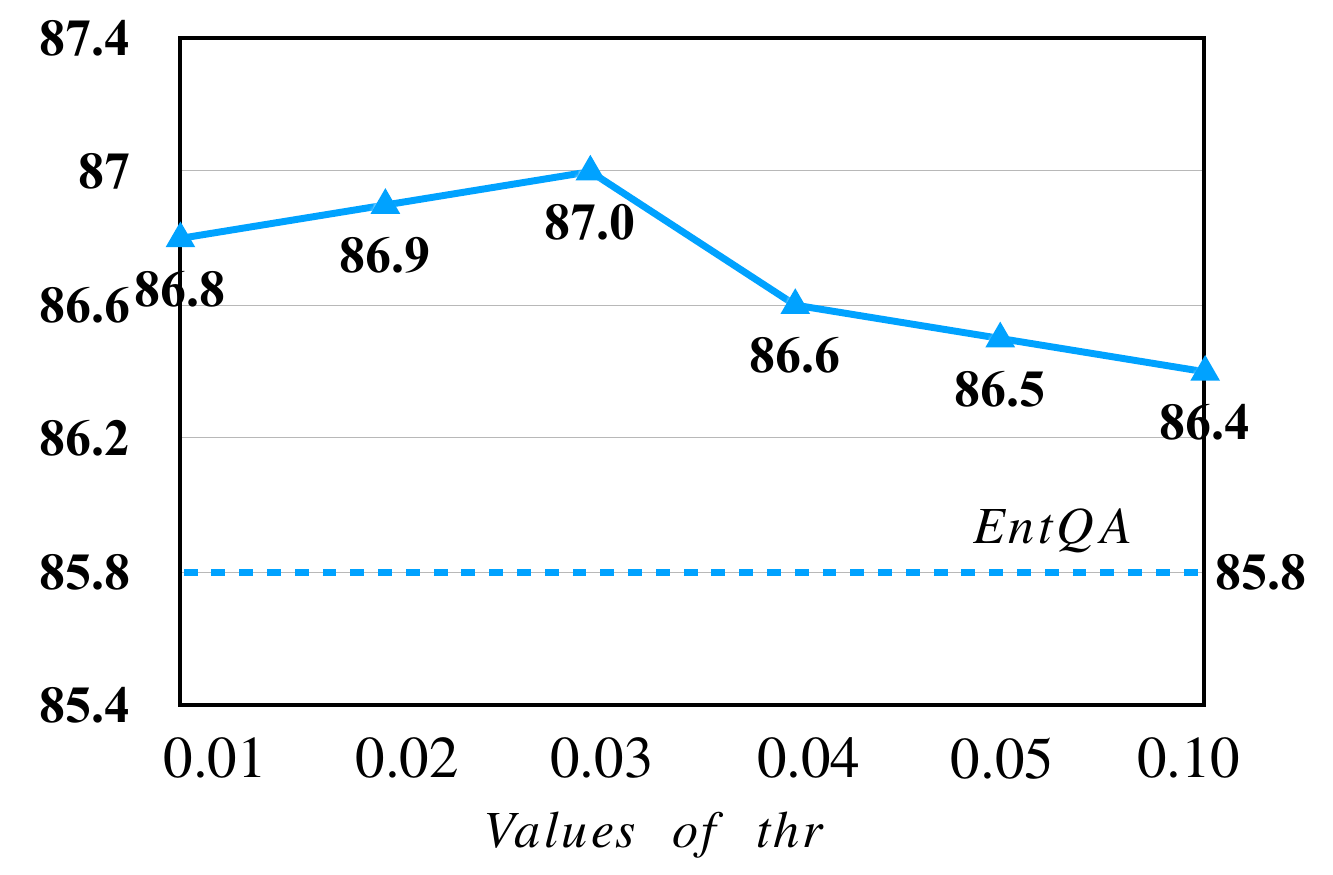} 
} 
\caption{
The F1 results of \modelName{} on AIDA-CoNLL using different parameters.
} 
\label{lambda_results} 
\end{figure} 

\subsubsection{Parameter Study of $K$ and $thr$}
As different amounts of candidate entities will affect \modelName{}'s performance, it is essential for us to study the effects of parameter $K$ in our designed retriever module. Figure~\ref{Parameters_K_Figure} presents the performance change of \modelName{} as choosing different values of $K$. We see that as the value of $K$ increases, the performance of the \modelName{} shows a trend of first increasing and then decreasing. This phenomenon is in line with the intuition and our design, because $K$ represents the number of candidate entities sent to the reader. If the value of $K$ is too large, it will cause more noise entities to be sent to the reader, thus damaging the reader's performance. However, in fact, choosing an excessively large $K$ value itself will not bring much gain to \modelName{}, and will even greatly increase the time spent by the process of retrieving entities. 
Therefore, choosing an appropriate value of $K$ can obtain competitive performance, after all, \modelName{} performs better than EntQA at all $K$ in Figure~\ref{Parameters_K_Figure}.

In Section~\ref{sec:reader}, we design to automatically filter predicted combinations of span and entity in the reader. As a key parameter, we carry out the parameter study to verify the insensitivity of \modelName{} to $thr$.  From Figure~\ref{Parameters_thr_Figure}, we see that the performance of \modelName{} is not very sensitive to the specific values when $thr$ is within a reasonable range. As $thr$ changes, the F1 score fluctuates slightly in the range greater than 85.0. Therefore, the performance of \modelName{} is robust to the choice of $thr$. And it can be seen that the performance of \modelName{} is always better than that of EntQA with the change of $thr$ value.

\begin{table}
\small
\centering
\begin{tabular}{p{1.5cm}p{10.0cm}}
\toprule
\textbf{Input 1:} & \textcolor{orange}{pakistan}, who arrive next week, are the third team in the triangular \textcolor{orange}{world series}  \\
\textbf{Gold:} & \textcolor{blue}{[0, 0, “Pakistan national cricket team”], [12, 13, “World Series Cricket”]} \\
\textbf{EntQA:} & [5, 5, “Pakistan national cricket team”] \\
\textbf{\modelName{}:} & [5, 5, “Pakistan national cricket team”], [12, 13, “World Series Cricket”]\\
\midrule

\textbf{Input 2:} & were not optimistic of a peaceful festive season in \textcolor{orange}{kwazulu-natal} \\
\textbf{Gold:} & \textcolor{blue}{[11, 15, “KwaZulu-Natal”]} \\
\textbf{EntQA:} & [11, 15, “KwaZulu-Natal”], \textcolor{red}{[11, 15, “KwaZulu”]} \\
\textbf{\modelName{}:} & [11, 15, “KwaZulu-Natal”] \\

\bottomrule
\end{tabular}
\caption{Examples of EntQA and \modelName{}. 
We mark the \textcolor{red}{span of mention}/\textcolor{blue}{golden entity}/\textcolor{orange}{wrong results}. For the EL task, the golden information includes the starting and ending positions of the span and the specific entity. }
\label{Case_Studies}
\end{table}

\subsection{Case Study}
Table~\ref{Case_Studies} illustrates the comparisons between the cases of EntQA and \modelName{}. In the first case, EntQA does not recognize that “world series” in the sentence is a mention of an entity, while \modelName{} does. We think this is because the retrieval results of \modelName{} are more diverse than that of EntQA because we leverage two kinds of representations for candidate retrieval. 
In addition, more interestingly, we find that \modelName{} is better than EntQA when dealing with nested entities, as shown in the second example. Because EntQA only uses the overall sentence representation of the $\text{[CLS]}$ for retrieval, it cannot perceive the specific position of the span, which leads the model to think that there may be two mentions in the sentence, namely “kwazulu” and “kwazulu-natal”. But \modelName{} knows the specific location of the span when retrieving, that is, it knows that [11, 15] is a span, so it can avoid selecting “kwazulu”. Therefore, the second example reflects the importance of the span information predicted by the reader for the retriever.

\section{Related Work}
\label{Sec:Related_Work}
End-to-End Entity Linking (EL) is a realistic yet challenging task and is the general form of EL~\citep{9560019, zhang-li-etal-2022-hosmel, DBLP:conf/cikm/JokoH22}. Early works divide the end-to-end EL task into two subtasks, namely Mention Detection (MD)/NER and Entity Disambiguation (ED)~\citep{DBLP:conf/cikm/SilY13}, and study the joint learning of these two subtasks to improve EL performance~\citep{luo-etal-2015-joint, nguyen-etal-2016-j, martins-etal-2019-joint}. \citep{kolitsas-etal-2018-end} develop the first neural end-to-end EL system that considers all potential mentions and calculates contextual similarity scores of candidate entities. 
Recently, researchers have gradually become enthusiastic about using paradigms of other tasks to improve the end-to-end EL performance~\citep{wu-etal-2020-scalable, de-cao-etal-2021-highly, lai-etal-2022-improving, cho2022unsupervised, DBLP:journals/tkde/RanSGLWJ23}.

GENRE~\citep{DBLP:conf/iclr/CaoI0P21} is an autoregressive model for end-to-end EL. It retrieves entities by generating entity names in an autoregressive mechanism~\citep{DBLP:journals/corr/abs-2112-11739}. ReFinED~\citep{ayoola-etal-2022-refined} separately address the EL task into three subtasks, namely MD, fine-grained entity typing, and ED, thereby enhancing EL with the help of fine-grained entity categories and descriptions. Considering the long-term dilemma of previous works performing MD before ED, that is these methods require models to accurately extract mentions without entity information, EntQA~\citep{DBLP:conf/iclr/ZhangHS22} of the retriever-reader paradigm is proposed to solve ED before MD by the way of inverted Open-Domain Question Answering. Thanks to its more natural design, EntQA has become a strong baseline that cannot be bypassed on the end-to-end EL task. Besides, the success of EntQA also indicates the advantages of the retriever-reader structure for end-to-end EL. Our work aims to propose a novel EL model in which the retriever and reader are more interactive to facilitate the advancement of the retriever-reader paradigm on end-to-end entity linking.

\section{Conclusion}
\label{Sec:Conclusion}
In this paper, we introduce to promote EL by facilitating the advancement of the retriever-reader paradigm. We design \modelName{}, a bidirectional end-to-end learning framework that enables sufficient retriever-reader interaction. Extensive experiments and analyses show the effectiveness of \modelName{}. In the future, we think it is a promising direction to apply our idea of enhancing the retriever-reader interaction to other related tasks. Besides, our practice of using span representations to assist retrieval is a valuable exploration for dense entity retrieval.

\section*{Acknowledgments}
This research is supported by National Natural Science Foundation of China (Grant No.62276154), Research Center for Computer Network (Shenzhen) Ministry of Education, Beijing Academy of Artificial Intelligence (BAAI), the Natural Science Foundation of Guangdong Province (Grant No. 2023A1515012914), Basic Research Fund of Shenzhen City (Grant No. JCYJ20210324120012033 and JSGG20210802154402007), Shenzhen Science and Technology Program (Grant No. WDZC20231128091437002), the Major Key Project of PCL for Experiments and Applications (PCL2021A06), and Overseas Cooperation Research Fund of Tsinghua Shenzhen International Graduate School (HW2021008).

\section*{Declaration of competing interest}
The authors declare that they have no known competing financial interests or personal relationships that could have appeared to influence the work reported in this paper.

\section*{CRediT authorship contribution statement}
Yinghui Li: Conceptualization,  Software, Formal analysis, Investigation, Methodology, Writing - original draft.

Yong Jiang: Conceptualization,  Formal analysis,  Methodology, Writing - original draft.

Yangning Li: Formal analysis, Validation, Visualization, Writing - original draft.

Xinyu Lu: Investigation, Data curation, Writing - review \& editing

Penjun Xie: Resources, Writing - review \& editing

Ying Shen: Writing - review \& editing

Hai-Tao Zheng: Funding acquisition, Supervision, Writing - review \& editing

\section*{Data availability}
The data used in our work will be made available on request.




\bibliographystyle{elsarticle-num-names} 

\bibliography{IEEEabrv}






\end{document}